\documentclass[conference]{IEEEtran}
\IEEEoverridecommandlockouts
\usepackage{cite}
\usepackage{amsmath,amssymb,amsfonts}
\usepackage{algorithmic}
\usepackage{graphicx}
\usepackage{textcomp}
\usepackage{xcolor}
\usepackage{hyperref}

\def\BibTeX{{\rm B\kern-.05em{\sc i\kern-.025em b}\kern-.08em
    T\kern-.1667em\lower.7ex\hbox{E}\kern-.125emX}}
\begin{document}
\title{Trustworthy Speech-Based Autism Screening under Class Imbalance with Uncertainty-Aware HuBERT Ensembles\\
}
\author{\IEEEauthorblockN{1\textsuperscript{st} Nihel Loubna Ghennani}
\IEEEauthorblockA{\textit{Department of Computer Science} \\
\textit{University Abu Bekr Belkaid}\\
\textit{Engineering Biomédical laboratory (GBM)}\\
\textit{IRIT, University of Toulouse}\\
Tlemcen, Algeria \\
nihelloubna.ghennani@univ-tlemcen.dz }
\and
\IEEEauthorblockN{2\textsuperscript{nd} Mourtada Benazzouz}
\IEEEauthorblockA{\textit{Department of Computer Science} \\
\textit{University Abu Bekr Belkaid}\\
\textit{Engineering Biomédical laboratory (GBM)}\\
Tlemcen, Algeria \\
mourtada.benazzouz@univ-tlemcen.dz}
\and
\IEEEauthorblockN{3\textsuperscript{rd} Julie Mauclair}
\IEEEauthorblockA{\textit{Department of Signal Processing} \\
\textit{IRIT, University of Toulouse}\\
Toulouse, France \\
julie.mauclair@irit.fr}
}
\maketitle
\begin{abstract}
Early screening of Autism Spectrum Disorder (ASD) is critical for timely clinical intervention, although establishing reliable speech based screening systems remains problematic due to limited clinical data, class imbalance, and the need for reliable uncertainty estimation. We propose an uncertainty-aware ASD screening framework based on self-supervised HuBERT representations, combining Leave-One-Subject-Out (LOSO) cross-validation, balanced k-bagging, and Monte Carlo (MC) Dropout. Multiple HuBERT models are fine-tuned on balanced subsets, with predictions combined through probability-based ensemble averaging and aggregated at the subject level. Experiments on the Child Pathological Speech Database (CPSD) show strong performance, achieving 86.30\% balanced accuracy and 93.74\% ROC-AUC at the utterance level, and 95.04\% and 96.30\%, respectively, at the subject level. Furthermore, predictive entropy effectively reflects classification reliability: incorrectly classified segments show higher entropy than correctly classified ones (0.4605 $\pm$ 0.2314 vs. 0.1717 $\pm$ 0.1830), with a significant correlation with errors ($r=0.3981$, $p=7.38\times10^{-170}$) and an error-detection ROC-AUC of 0.8359. These results demonstrate the potential of the proposed framework as a robust and uncertainty-aware computer-aided tool for ASD screening.

\end{abstract}
\section{Introduction}
\label{sec:intro}
Autism Spectrum Disorder (ASD) is a neurodevelopmental condition characterized by persistent deficits in social communication together with restricted and repetitive patterns of behavior \cite{b1,b2}. Since there is currently no cure for ASD, early identification and timely intervention are essential for improving developmental outcomes and helping children acquire the skills needed for daily life \cite{b3}. Among the behavioral manifestations of ASD, atypical speech prosody has emerged as a promising early biomarker \cite{b4}, motivating the development of automated speech-based diagnostic systems capable of providing objective and reproducible clinical support \cite{b5}.
The conventional approaches for ASD screening that make use of speech mainly depend on hand-crafted acoustic feature sets, namely pitch, jitter, shimmer, spectral features, and MFCCs \cite{b6,b7}. Recent advancements in the field of SSL models like wav2vec 2.0 \cite{b8} and HuBERT \cite{b9} have led to significant improvements in learning representations for speech through contextual embedding of large unlabeled speech corpora. These foundation models have demonstrated 
superior performance across numerous speech analysis tasks and offer considerable 
potential for clinical speech assessment \cite{b10}.

Despite these improvements, there are still several challenges limit the reliable deployment of SSL models for ASD screening. Moreover, clinical speech data is generally small and highly imbalanced, making it susceptible to overfitting by the deep learning model as well as favoring the dominant class. Additionally, the conventional method of fine-tuning relies on only one deterministic model, and thus cannot provide any confidence levels, making the predictions less interpretable and reliable.

To address these limitations, we propose an uncertainty-aware HuBERT ensemble framework for speech-based ASD screening. The main contributions of this work are summarized as follows:
\begin{itemize}
\item We propose a HuBERT-based ensemble framework that combines self-supervised speech representations, balanced \textit{K}-bagging, and Monte Carlo Dropout to address class imbalance and estimate predictive uncertainty for robust ASD screening.
\item We validate the proposed framework through an ablation study at both the utterance and subject levels under the LOSO protocol, evaluating discrimination, calibration, and uncertainty to comprehensively assess its performance and reliability.
\end{itemize}
The rest of this paper is organized as follows. Section~\ref{sec:method}  describes the methodology that we propose. Section~\ref{sec:expset} presents the experimental set-up. The experimental results are described in Section~\ref{sec:expres}. The results are discussed in Section~\ref{sec:dis}. Section~\ref{sec:con} summarizes the findings and highlights future research.

\section{Proposed Methodology}
\label{sec:method}
\begin{figure*}[!t]
\centering
\includegraphics[width=\textwidth]{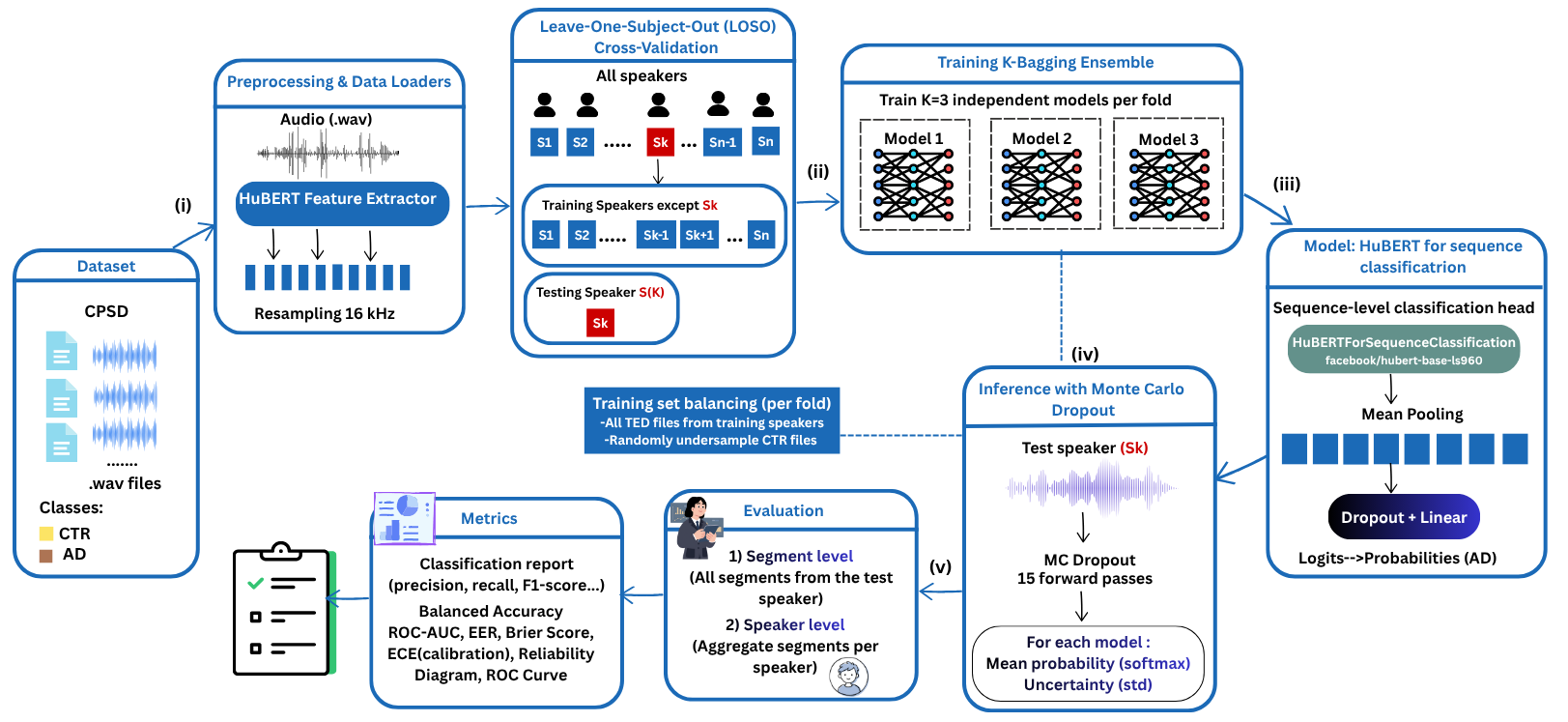}
\caption{Overview of the proposed uncertainty-aware HuBERT ensemble framework for speech-based ASD screening.}
\label{fig:methodology}
\end{figure*}

The complete pipeline of the proposed speech based ASD screening framework is illustrated in Fig.~\ref{fig:methodology}. The proposed methodology consists of five main stages: (i) audio preprocessing and data loading, (ii) Leave-One-Subject-Out (LOSO) cross-validation with balanced bagging, (iii) self-supervised feature extraction and classification using HuBERT, (iv) uncertainty-aware ensemble prediction with Monte Carlo (MC) Dropout, and (v) subject-level probability aggregation for clinical decision support.

\noindent\textbf{(i) Preprocessing and Data Loading}

All speech recordings are resampled to 16~kHz, which is the input format expected by the pre-trained HuBERT-Base model . The audio signals are then processed by the HuBERT feature extractor to get the input representations needed for model fine-tuning and inference.

\noindent\textbf{(ii) Cross-Validation and Balanced Bagging}

A Leave-One-Subject-Out (LOSO) cross-validation protocol is adopted to ensure subject-independent evaluation. In each fold, all utterances of one participant are kept for testing, and the other participants are used for training, preventing speaker dependent data leakage.
To address the class imbalance between the Control (CTR) and Autistic Disorder (AD) groups, a balanced bagging strategy is applied on each LOSO fold. All AD samples are retained, and the majority class is randomly under-sampled to obtain $K$ balanced training subsets,
\[
\{D_1,D_2,\ldots,D_K\},
\]

where each subset contains all AD utterances and an equal number of randomly CTR utterances. An independent HuBERT model is fine-tuned on each subset, forming an ensemble of $K$ classifiers.

\noindent\textbf{(iii) HuBERT-Based Speech Classification}

Given an input speech waveform, the pre-trained HuBERT Base encoder extracts contextualized speech representations, which are aggregated using temporal pooling to get a fixed dimensional utterance embedding,
\begin{equation}
\mathbf{z}=\frac{1}{T}\sum_{t=1}^{T}H_t,
\label{eq:pooling}
\end{equation}

where $H_t$ denotes the hidden representation at time step $t$ and $T$ is the number of frames. The pooled embedding is then passed to the sequence classification head to estimate the posterior probabilities of the CTR and AD classes.

\noindent\textbf{(iv) Monte Carlo Dropout and Ensemble Fusion}

To improve prediction reliability and quantify model uncertainty, Monte Carlo (MC) Dropout \cite{b11} is employed during inference. During testing, dropout layers remain active and each HuBERT model performs $N$ stochastic forward passes. The predictive probability of the $k$-th model is obtained by averaging the stochastic predictions,

\begin{equation}
p_k=
\frac{1}{N}
\sum_{n=1}^{N}
M_k^{(n)}(\mathbf{z}),
\label{eq:mc_dropout}
\end{equation}

The final ensemble probability is computed by soft voting over the $K$ HuBERT models,

\begin{equation}
P(y=1|\mathbf{z})
=
\frac{1}{K}
\sum_{k=1}^{K}
p_k,
\label{eq:final_prob}
\end{equation}

Model uncertainty is estimated from the standard deviation of the MC Dropout predictions, while predictive entropy is computed from the final ensemble probability.

\noindent\textbf{(v) Subject-Level Decision Aggregation}

The final subject-level prediction is obtained by averaging the ensemble probabilities of all utterances belonging to the same participant:
\begin{equation}
P_s=
\frac{1}{L_s}
\sum_{i=1}^{L_s}
P_i,
\label{eq:subject_prediction}
\end{equation}
where $L_s$ is the number of utterances of subject $s$ and $P_i$ is the ensemble probability of the $i$-th utterance.
\section{Experimental Setup}
\label{sec:expset}
\subsection{Dataset}
The proposed framework is evaluated on the Child Pathological Speech Database (CPSD) \cite{b12, b13}, which contains speech recordings from 35 children with neurodevelopmental disorders, including 12 with Autistic Disorder (AD), 10 with Pervasive Developmental Disorder-Not Otherwise Specified (PDD-NOS), and 13 with Specific Language Impairment (SLI), along with 70 typically developing controls (CTR). The recordings were collected during a semi-structured storytelling task using \textit{Frog, Where Are You?}, comprising approximately 10 hours of speech. After segmentation and manual quality control, this study focuses on distinguishing children with AD (disorder group) from typically developing controls (CTR).

\subsection{Implementation Details}
The proposed framework is implemented in PyTorch using the pre-trained
\texttt{hubert-base-ls960} model as the backbone network. During fine-tuning,
the convolutional feature encoder is frozen, while the Transformer encoder and
the sequence classification head are optimized using the AdamW optimizer with a
learning rate of $1\times10^{-5}$. Training is performed for 3 epochs with a
batch size of 4. For each LOSO fold, balanced bagging is employed to construct an 
ensemble of $K=3$ independently fine-tuned HuBERT models, each trained on a different
balanced subset of the training data. During inference, Monte Carlo (MC)
Dropout remains active, and $N=15$ stochastic forward passes are performed for
each model to estimate predictive uncertainty. The final prediction is obtained
by averaging the predictive probabilities of the ensemble members through soft
voting.
\subsection{Evaluation Metrics}
To comprehensively evaluate the proposed approach under the class-imbalanced dataset, performance is measured at both the utterance (segment) and subject levels utilizing three complimentary perspectives: classification performance, probabilistic calibration, and predictive uncertainty.
\subsubsection{Classification and Discrimination Metrics}
Considering the imbalance between the Control (CTR) and AD classes, the primary performance metric is the Balanced Accuracy \cite{b14} ($\mathrm{BACC}$), defined as

\begin{equation}
\mathrm{BACC} =
\frac{1}{2}\left(
\frac{\mathrm{TP}}{\mathrm{TP}+\mathrm{FN}}
+
\frac{\mathrm{TN}}{\mathrm{TN}+\mathrm{FP}}
\right),
\label{eq:bacc}
\end{equation}
where $\mathrm{TP}$, $\mathrm{TN}$, $\mathrm{FP}$, and $\mathrm{FN}$ denote the numbers of true positives, true negatives, false positives, and false negatives, respectively.
The threshold-independent discriminative capability of the proposed framework is further assessed using the Receiver Operating Characteristic Area Under the Curve (ROC-AUC), and the Equal Error Rate (EER). At the subject level, utterance-level probabilities are averaged for each participant, and the resulting predictions are evaluated using Sensitivity, Specificity, Positive Predictive Value (PPV), and Negative Predictive Value (NPV).
\subsubsection{Probabilistic Calibration}
To assess the reliability of the predicted posterior probabilities, calibration is evaluated using the Brier Score \cite{b15} and the Expected Calibration Error (ECE) \cite{b16}. The Brier Score is defined as
\begin{equation}
\mathrm{Brier} =
\frac{1}{M}\sum_{i=1}^{M}
\left(P(y_i \mid \mathbf{z}_i) - y_i\right)^2,
\label{eq:brier}
\end{equation}
where $M$ is the number of test samples. The ECE \cite{b17} measures the weighted discrepancy between the predicted confidence and the observed accuracy across $B=10$ 
equally spaced confidence bins:
\begin{equation}
\mathrm{ECE} =
\sum_{b=1}^{B}
\frac{|I_b|}{M}
\left|
\mathrm{acc}(I_b) - \mathrm{conf}(I_b)
\right|,
\label{eq:ece}
\end{equation}
Reliability diagram \cite{b18} is additionally reported to provide a visual assessment of model calibration.

\subsubsection{Predictive Uncertainty}

Predictive uncertainty is quantified using predictive entropy, where larger entropy values indicate lower prediction confidence. The standard deviation of the Monte Carlo Dropout \cite{b11} predictions is also reported as an estimate of epistemic uncertainty. To assess the relationship between uncertainty and prediction reliability, predictive entropy is further evaluated using its correlation with classification errors and its ability to discriminate between correct and incorrect predictions through ROC-AUC.
\section{Experimental Results}
\label{sec:expres}
This section presents the experimental results of the proposed HuBERT ensemble framework using the Leave-One-Subject-Out (LOSO) protocol. Performance was evaluated at both utterance and subject levels, which provide complementary perspectives: utterance-level evaluation assesses individual speech segments, while subject-level evaluation aggregates segment probabilities to produce one prediction per participant. Four configurations were progressively evaluated: A0 (baseline HuBERT), A1 (HuBERT + balanced bagging), A2 (HuBERT + balanced bagging + ensemble learning), and A3 (the complete framework with MC dropout).

The utterance-level classification results obtained with these four configurations are presented in Table~\ref{tab:utterance_results}, while the corresponding subject-level results are presented in Table~\ref{tab:subject_results}

\begin{table*}[!t]
\centering
\caption{Ablation study at the utterance (segment) level.}
\label{tab:utterance_results}
\begin{tabular*}{\textwidth}{@{\extracolsep{\fill}}lcccccccc@{}}
\hline
\textbf{Method} &
\textbf{Bal. Acc.} $\uparrow$ &
\textbf{Recall(AD)} $\uparrow$ &
\textbf{F1(AD)} $\uparrow$ &
\textbf{ROC-AUC} $\uparrow$ &
\textbf{PR-AUC} $\uparrow$ &
\textbf{EER} $\downarrow$ &
\textbf{Brier} $\downarrow$ &
\textbf{ECE} $\downarrow$ \\
\hline

A0
& 0.7583 & 0.55 & 0.63 & 0.9195 & 0.7063 & 0.1708 & 0.0805 & 0.0730 \\

A1
& 0.8349 & 0.74 & 0.68 & 0.9184 & 0.7384 & 0.1561 & 0.0847 & 0.0694 \\

A2
& 0.8302 & 0.72 & 0.69 & 0.9331 & 0.7554 & 0.1468 & 0.0721 & 0.0508 \\

\textbf{A3}
& \textbf{0.8630} & \textbf{0.79} & \textbf{0.73} &
\textbf{0.9374} & \textbf{0.7594} & \textbf{0.1298} &
\textbf{0.0675} & 0.0515 \\

\hline
\end{tabular*}
\end{table*}

\begin{table*}[!t]
\centering
\caption{Progressive ablation study at the subject level under LOSO cross-validation.}
\label{tab:subject_results}
\begin{tabular*}{\textwidth}{@{\extracolsep{\fill}}lcccccccc@{}}
\hline
\textbf{Method} &
\textbf{Bal. Acc.} $\uparrow$ &
\textbf{Sens.} $\uparrow$ &
\textbf{Spec.} $\uparrow$ &
\textbf{PPV} $\uparrow$ &
\textbf{NPV} $\uparrow$ &
\textbf{ROC-AUC} $\uparrow$ &
\textbf{EER} $\downarrow$ &
\textbf{Brier} $\downarrow$ \\
\hline

A0
& 0.7837 & 0.5833 & 0.9841 & 0.8750 & 0.9254
& 0.9735 & 0.0833 & 0.0520 \\

A1
& 0.9425 & \textbf{0.9167} & 0.9683 & 0.8462 & \textbf{0.9839}
& \textbf{0.9881} & \textbf{0.0476} & \textbf{0.0400} \\

A2
& 0.9087 & 0.8333 & \textbf{0.9841} & 0.9091 & 0.9688
& 0.9312 & 0.0833 & 0.0439 \\

A3
& \textbf{0.9504} & \textbf{0.9167} & 0.9841 & \textbf{0.9167} & \textbf{0.9841}
& 0.9630 & 0.0833 & 0.0400 \\

\hline
\end{tabular*}
\end{table*}

\begin{figure}[!t]
\centering
\includegraphics[width=\columnwidth]{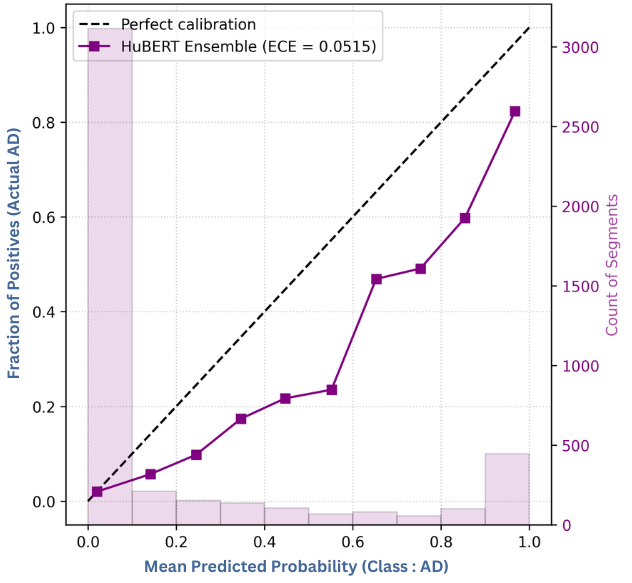}
\caption{Utterance-level reliability diagram of the proposed HuBERT ensemble.}
\label{fig:reliability}
\end{figure}

\begin{figure}[!t]
\centering
\includegraphics[width=\columnwidth]{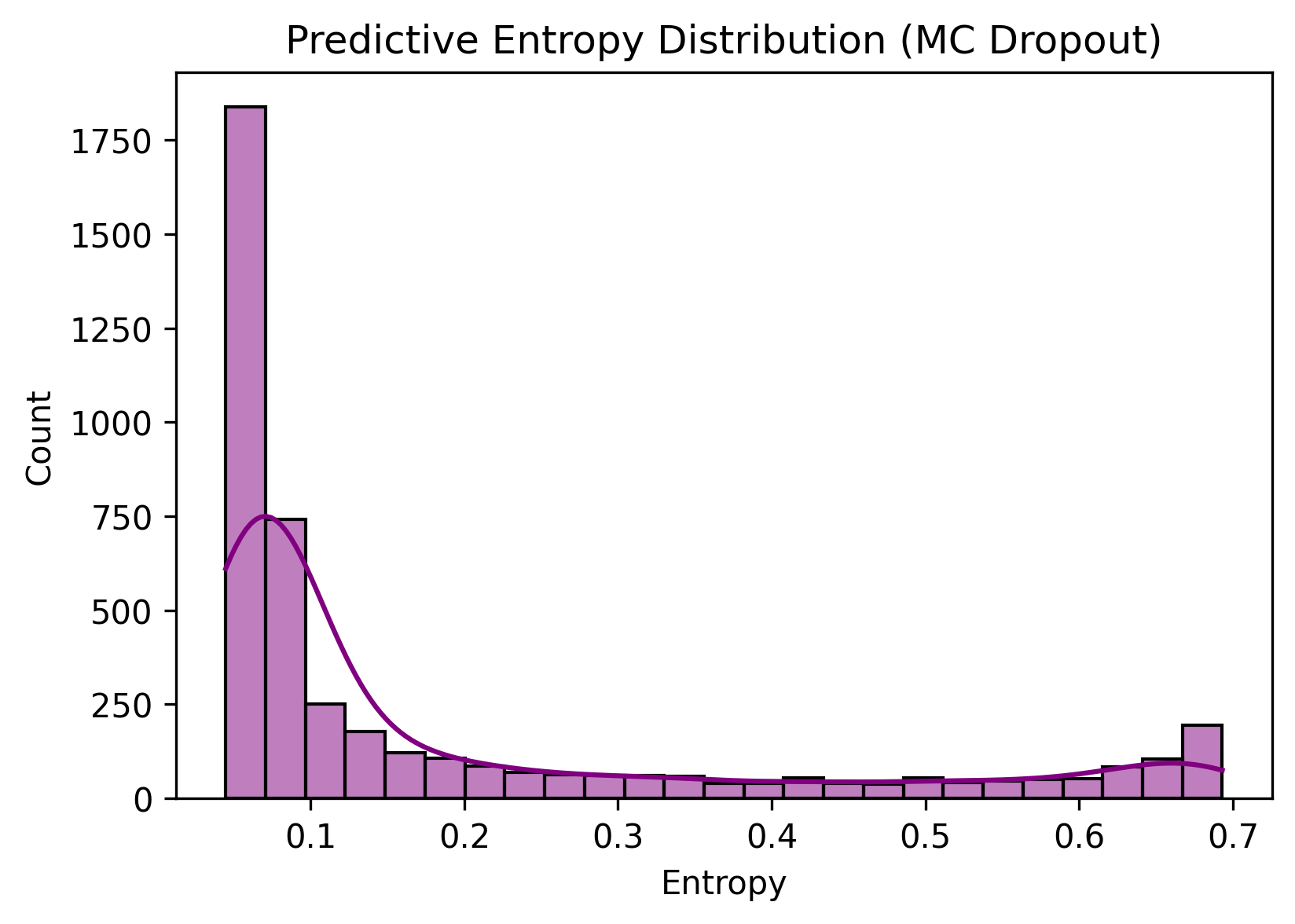}
\caption{Distribution of predictive entropy obtained using Monte Carlo dropout.}
\label{fig:entropy}
\end{figure}

Probability calibration is assessed using the reliability diagram shown in Fig.~\ref{fig:reliability}. Finally, predictive uncertainty is characterized through the distribution of predictive entropy obtained using MC dropout during inference, as illustrated in Fig~\ref{fig:entropy}.

Predictive entropy was further analyzed in relation to actual classification errors. Table~\ref{tab:entropy_error} summarizes the entropy differences between correct and incorrect predictions and their association with classification errors.
\begin{table}[!t]
\centering
\caption{Relationship between predictive entropy and utterance-level classification errors.}
\label{tab:entropy_error}
\begin{tabular}{lc}
\hline
\textbf{Measure} & \textbf{Value} \\
\hline
Correct predictions & $0.1717 \pm 0.1830$ \\
Incorrect predictions & $0.4605 \pm 0.2314$ \\
Entropy gap & $0.2888$ \\
Point-biserial correlation ($r$) & $0.3981$ \\
$p$-value & $7.38\times10^{-170}$ \\
Error-detection ROC-AUC & $0.8359$ \\
\hline
\end{tabular}
\end{table}


\section{Discussion}
\label{sec:dis}
The proposed uncertainty-aware HuBERT ensemble achieved strong discrimination performance for ASD screening at both the utterance and subject levels. At the utterance level, the complete configuration (A3), combining HuBERT, balanced bagging, ensemble learning, and MC dropout, achieved the highest balanced accuracy (0.8630), AD recall (0.79), F1-score (0.73), ROC-AUC (0.9374), and PR-AUC (0.7594), together with the lowest EER (0.1298) and Brier score (0.0675). These results demonstrate the effectiveness of the complete configuration for classifying individual speech segments. At the subject level, A3 achieved a balanced accuracy of 0.9504, sensitivity of 0.9167, specificity of 0.9841, PPV of 0.9167, and NPV of 0.9841. The subject-level and utterance-level results should be interpreted as complementary evaluations rather than as one improving upon the other, since they correspond to different prediction units. Utterance-level metrics characterize the classification of individual speech segments, whereas subject-level metrics characterize the final participant-level decision after aggregating segment-level posterior probabilities.

The utterance-level reliability diagram in Fig.~\ref{fig:reliability} provides a visual assessment of the calibration of the proposed HuBERT ensemble. The calibration curve lies predominantly below the perfect-calibration diagonal, particularly in the higher predicted-probability range, indicating a tendency toward overconfident predictions. Thus, although the model exhibits strong discrimination, its predicted probabilities are not perfectly aligned with the observed outcome frequencies. This observation is consistent with the ECE of 0.0515 and Brier score of 0.0675 obtained by A3, which indicate a relatively small but non-negligible calibration error. The reliability analysis therefore provides information that is complementary to the discrimination metrics by characterizing the quality of the predicted probabilities.

The predictive entropy distribution obtained using MC dropout is illustrated in Fig.~\ref{fig:entropy}. Most utterances exhibit relatively low entropy, indicating that the model produces confident predictions for a large proportion of the segments, while a smaller portion presents higher uncertainty. The relationship between this uncertainty and actual classification outcomes is further quantified in Table~\ref{tab:entropy_error}. Incorrectly classified segments exhibit substantially higher mean entropy than correctly classified segments (0.4605 $\pm$ 0.2314 vs. 0.1717 $\pm$ 0.1830), resulting in an entropy gap of 0.2888. Moreover, predictive entropy shows a positive and statistically significant association with classification errors ($r=0.3981$, $p=7.38\times10^{-170}$). The error-detection ROC-AUC of 0.8359 further indicates that entropy provides useful discrimination between incorrect and correct predictions. These results establish a direct relationship between the uncertainty estimated by MC dropout and prediction errors.

Overall, the results demonstrate strong and complementary performance in classification, calibration, and uncertainty estimation. However, limitations include residual calibration errors, and evaluation on a single dataset. Further validation on diverse clinical cohorts is needed to confirm the model’s clinical utility.
\section{Conclusion}
\label{sec:con}
This work presente an uncertainty-aware HuBERT ensemble framework for speech-based Autism Spectrum Disorder (ASD) screening. the proposed approach addresses class imbalance while providing reliable uncertainty estimates. Experimental results demonstrated strong performance at both the utterance and subject levels, achieving a balanced accuracy 86.30\% and 95.04\% respectively. At the utterance level, the recall of Autistic Disorder (AD) increased from 55\% with the baseline, which was biased toward CTR as the majority class to 79\% with the full method, while sensitivity improved from 58.33\% to 91.67\% at subject level. The brier score which decrease from 8\% to 6\% at the  utterance level and from 5.2\% to 4\% at subject level.Moreover, predictive entropy was significantly associated with utterance-level classification errors ($r=0.3981$, $p=7.38\times10^{-170}$), with higher entropy observed for incorrect predictions. These results highlight the potential of the framework for computer aided ASD screening. Future work will focus on multimodal assessment, uncertainty-aware learning, Smartphone based screening, and expanding the system to additional disoder classes, including SLI and PDD-NOS .


\begin{thebibliography}{00}
\bibitem{b1} S. M. M. Hasan, M. P. Uddin, M. A. Mamun, M. I. Sharif, A. Ulhaq, and G. Krishnamoorthy, ``A machine learning framework for early-stage detection of autism spectrum disorders,'' \emph{IEEE Access}, vol. 11, pp. 15038--15057, 2023, doi: 10.1109/ACCESS.2022.3232490.

\bibitem{b2} M. Bala, M. H. Ali, M. S. Satu, K. F. Hasan, and M. A. Moni, ``Efficient machine learning models for early stage detection of autism spectrum disorder,'' \emph{Algorithms}, vol. 15, no. 5, Art. no. 166, 2022, doi: 10.3390/a15050166.

\bibitem{b3} N. L. Ghennani, J. Mauclair, and M. Benazzouz, ``Contributions de caractéristiques pour la détection de parole d’enfants autistes,'' in \emph{Proc. 36th Journées d’Études sur la Parole (JEP)}, 2026, pp. 72--81, doi: 10.21437/JEP.2026-8.

\bibitem{b4} H. Rakotomanana and G. Rouhafzay, ``A scoping review of AI-based approaches for detecting autism traits using voice and behavioral data,'' \emph{Bioengineering}, vol. 12, no. 11, Art. no. 1136, 2025, doi: 10.3390/bioengineering12111136.

\bibitem{b5} F. Briend, C. David, S. Silleresi, \emph{et al.}, ``Voice acoustics allow classifying autism spectrum disorder with high accuracy,'' \emph{Translational Psychiatry}, vol. 13, Art. no. 250, 2023, doi: 10.1038/s41398-023-02554-8.
\bibitem{b6}
``Multidisciplinary characterization of embarrassment through behavioral and acoustic modeling,''
\emph{Scientific Reports}, 2025, doi: 10.1038/s41598-025-94051-9.

\bibitem{b7}
D. M. Low, K. H. Bentley, and S. S. Ghosh, ``Automated assessment of psychiatric disorders using speech: A systematic review,'' \emph{Laryngoscope Investigative Otolaryngology}, vol. 5, no. 1, pp. 96--116, 2020.

\bibitem{b8}
A. Baevski, H. Zhou, A. Mohamed, and M. Auli, ``wav2vec 2.0: A framework for self-supervised learning of speech representations,'' in \emph{Advances in Neural Information Processing Systems (NeurIPS)}, vol. 33, 2020, pp. 12449--12460.

\bibitem{b9}
W.-N. Hsu, B. Bolte, Y.-H. H. Tsai, K. Lakhotia, R. Salakhutdinov, and A. Mohamed, ``HuBERT: Self-supervised speech representation learning by masked prediction of hidden units,'' \emph{IEEE/ACM Transactions on Audio, Speech, and Language Processing}, vol. 29, pp. 3451--3460, 2021.

\bibitem{b10}
A. Mohamed, H.-Y. Lee, L. B. Wang, R. N. Jakobsen, \emph{et al.}, ``Self-supervised speech representation learning: A review,'' \emph{IEEE Journal of Selected Topics in Signal Processing}, vol. 16, no. 6, pp. 1179--1210, 2022.

\bibitem{b11}
Y. Gal and Z. Ghahramani, ``Dropout as a Bayesian approximation: Representing model uncertainty in deep learning,'' \emph{arXiv preprint arXiv:1506.02142}, 2015, doi: 10.48550/arXiv.1506.02142.

\bibitem{b12}
J. Demouy, M. Plaza, J. Xavier, F. Ringeval, M. Chetouani, D. Périsse, D. Chauvin, S. Viaux, B. Golse, D. Cohen, and L. Robel, ``Differential language markers of pathology in Autism, Pervasive Developmental Disorder Not Otherwise Specified and Specific Language Impairment,'' \emph{Research in Autism Spectrum Disorders}, vol. 5, no. 4, pp. 1402--1412, Oct.--Dec. 2011, doi: \href{https://doi.org/10.1016/j.rasd.2011.01.026}{10.1016/j.rasd.2011.01.026}.

\bibitem{b13}
F. Ringeval, J. Demouy, G. Szaszák, M. Chetouani, L. Robel, J. Xavier, D. Cohen, and M. Plaza, ``Automatic intonation recognition for the prosodic assessment of language-impaired children,'' \emph{IEEE Transactions on Audio, Speech, and Language Processing}, vol. 19, no. 5, pp. 1328--1342, Jul. 2011, doi: \href{https://doi.org/10.1109/TASL.2010.2090147}{10.1109/TASL.2010.2090147}.

 \bibitem{b14}
H. S. Alsorory and M. Alshraideh, ``Boosting software fault prediction: Addressing class imbalance with enhanced ensemble learning,'' \emph{Applied Computational Intelligence and Soft Computing}, vol. 2024, Art. no. 2959582, 2024, doi: 10.1155/2024/2959582.

\bibitem{b15}
M. B. et al., ``Classifier calibration: A survey on how to assess and improve predicted class probabilities,'' \emph{Machine Learning}, 2024, doi: 10.1007/s10994-023-06336-7.

\bibitem{b16}
M. Pakdaman Naeini, G. F. Cooper, and M. Hauskrecht, ``Obtaining well calibrated probabilities using Bayesian binning,'' in \emph{Proc. AAAI Conf. Artif. Intell.}, vol. 29, no. 1, pp. 2901--2907, 2015, doi: \href{https://doi.org/10.1609/aaai.v29i1.9602}{10.1609/aaai.v29i1.9602}.

\bibitem{b17}
Y. Nie, X. Zhou, and M. Bansal, ``What can we learn from collective human opinions on natural language inference data?,'' in \emph{Proc. Conf. Empirical Methods in Natural Language Processing (EMNLP)}, Nov. 2020, pp. 9131--9143, doi: \href{https://doi.org/10.18653/v1/2020.emnlp-main.734}{10.18653/v1/2020.emnlp-main.734}.

\bibitem{b18}
M. Pavlovic, ``Understanding model calibration: A gentle introduction and visual exploration of calibration and the expected calibration error (ECE),'' \emph{ICLR Blogposts}, Apr. 2025. doi: \href{https://doi.org/10.48550/arXiv.2501.19047}{10.48550/arXiv.2501.19047}.
\end{thebibliography}
\end{document}